\documentclass{article}
\usepackage{ijcai23}

\usepackage{times}
\usepackage{soul}
\usepackage{url}
\usepackage[hidelinks]{hyperref}
\usepackage[utf8]{inputenc}
\usepackage[small]{caption}
\usepackage{graphicx}
\usepackage{amsmath}
\usepackage{amsthm}
\usepackage{booktabs}
\usepackage{algorithm}
\usepackage[switch]{lineno}
\usepackage{enumitem}
\usepackage{amsfonts}
\usepackage{algpseudocode}
\usepackage{array}
\usepackage{multirow}
\usepackage{verbatim}
\usepackage{multicol}
\usepackage{multirow}

\title{Learning Compact Neural Networks with Deep Overparameterised Multitask Learning}

\author{
Shen Ren$^1$
\and
Haosen Shi$^2$ \\
\affiliations
$^1$Continental Automotive Singapore\\
$^2$Continental-NTU Corporate Lab, Nanyang Technological University
\emails
shen.ren@continental-corporation.com,
haosen.shi@ntu.edu.sg
}

\begin{document}

\maketitle

\begin{abstract}
Compact neural network offers many benefits for real-world applications. However, it is usually challenging to train the compact neural networks with small parameter sizes and low computational costs to achieve the same or better model performance compared to more complex and powerful architecture. This is particularly true for multitask learning, with different tasks competing for resources. We present a simple, efficient and effective multitask learning overparameterisation neural network design by overparameterising the model architecture in training and sharing the overparameterised model parameters more effectively across tasks, for better optimisation and generalisation. Experiments on two challenging multitask datasets (NYUv2 and COCO) demonstrate the effectiveness of the proposed method across various convolutional networks and parameter sizes.
\end{abstract}

\section{Introduction}

Deep Multi-task Learning (MTL) techniques are widely applied to real-world embedded computer vision applications. A deep MTL model explores and exploits the synergies among multiple tasks to be learned simultaneously to improve the joint performance, as well as to reduce inference time and computational costs. However, designing efficient MTL models on budgeted devices poses two major challenges. First, the model design needs to be efficient to stay compact for meeting the computational budget constraints. Second, the model needs to be effective on resource sharing among multiple tasks learned simultaneously to avoid resource competition. 

Motivated by previous research on deep linear network \cite{saxe2013exact} that even the input-output map can be rewritten as a shallow network, it nevertheless demonstrates highly nonlinear training dynamics and can help to accelerate optimisation \cite{arora2018optimization} and improve generalisation \cite{advani2020high}. To tackle the aforementioned challenges, we propose an overparameterised MTL method by initialising the parameters of each shared neural network layer as the product of multiple matrices following the spatial Singular Vector Decomposition (SVD)\cite{jaderberg2014speeding}. The left and right singular vectors are trained with all task losses, and the diagonal matrices are trained using task-specific losses. Our design is mainly inspired by analytical studies on overparameterised networks for MTL \cite{lampinen2018analytic} that the training/test error dynamics depends on the time-evolving alignment of the network parameters to the singular vectors of the training data, and a quantifiable task alignment describing the transfer benefits among multiple tasks depends on the singular values and input feature subspace similarity matrix of the training data.

In this work, we follow the definition of overparameterisation~\cite{arora2018optimization}, referring to the replacement of neural network layers by operations of the compositions of multiple layers with more learnable parameters, but without adding additional expressiveness of the network. 
The contribution of this work can be summarised as follows.



\begin{itemize}[nolistsep]
    \item We propose an MTL neural network design with overparameterised training components and a compact inference architecture, applicable for embedded applications with limited computational budgets.
    \item We implement an iterative training strategy for the proposed design that is effective and efficient for the multitask computer vision dense prediction tasks, compared to the state-of-the-art.
\end{itemize}

\section{Methodology}

We replace the fully-connected layers and/or convolutional layers of modern neural networks with overparameterisation, and share the overparameterised parameters among different tasks, to achieve higher performance for reduced inference parameter size and computational cost.

Specifically, tensor decomposition is used for model expansion instead of model compression during training. The full-rank diagonal tensor is further expanded to be trained separately for each task, while the other tensors are shared among all tasks. During inference, the decomposed tensors are contracted back into a compact MTL architecture.



\subsection{Overparameterisation Mechanism}

\subsubsection{Fully-connected Layers}







For any shared layer of a deep MTL model, given a weight matrix $W$ that is shared among $t$ tasks, 
we directly factorise the weight matrix $W$ of the size $m \times n$ using SVD, similar to \cite{khodak2021initialization} and \cite{cao2020conv}, so that

\begin{equation}
  W := UMV
  \label{eq:multiply}
\end{equation}

$U$ is of size $m \times r$ and V is of size $r \times n$, and $M$ is a diagonal matrix of size $r \times r$ where the matrix is full rank $r\geq \min(m, n)$.

\subsubsection{Convolutional Layers}

A shared convolutional layer of a deep MTL model is parameterised as $W \in \mathbb{R}^{c_o \times c_i \times k \times k}$ to mathematically denote a 2-dimensional convolutional layer of kernel size $k \times k$ with $c_o$ output channels and $c_i$ input channels. 







Using tensor factorisation on the convolutional layer following the spatial SVD format, we first change the dimensions of the weight tensor into $c_o \times k \times k \times c_i$, and replace the original weight tensor into three tensors $U$, $M$ and $V$ using Eq. \eqref{eq:multiply}  with the size $c_o \times k \times r$, $r \times r$ and $r \times k \times c_i$ respectively, where $r\geq \min(c_o \times k, c_i \times k)$.  

\begin{figure}[h!]
  \centering
  \includegraphics[width=0.9\linewidth]{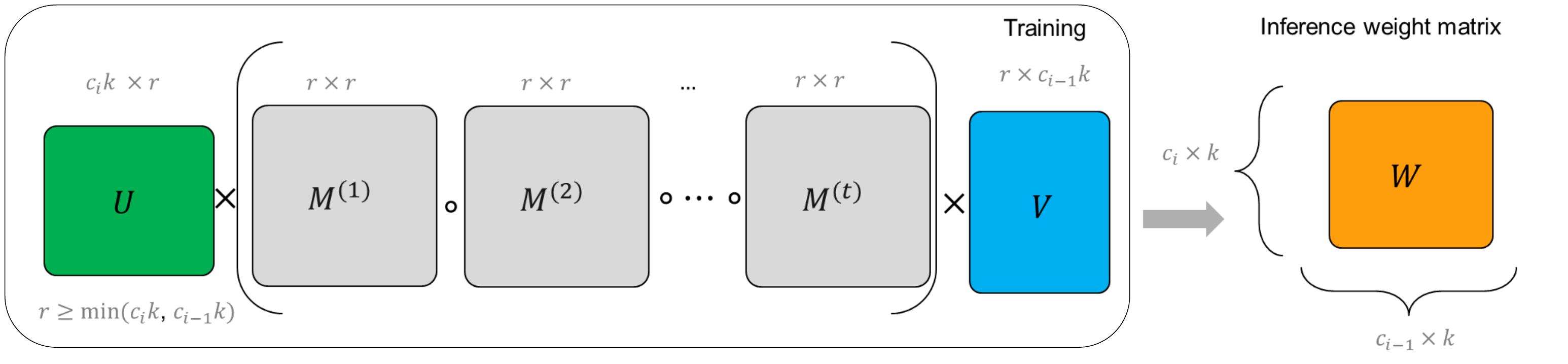}
  \caption{Overparameterisation and Parameter Sharing Mechanisms}
  \label{fig:fac}
\end{figure}


\subsection{Parameter Sharing Mechanism}

Considering each shared fully-connected or convolutional layer of an MTL model with the objective of learning tasks $a$ and $b$ together, after overparameterisation, the parameters $U$ and $V$ are shared across all tasks, and $ M^{(a)}, M^{(b)}$ are assigned as task-specific parameters for the corresponding tasks. The task-specific parameters $M^{(a)}$ and $M^{(b)}$ are learned as scaling factors in changing the scales of shared parameters $U$ and $V$ according to each individual task, to better align with the singular vectors of the training data. The product is cumulative, associative and distributive so that the sequence of the tasks will not take effect on the final product.


The designed sharing mechanism can be extended naturally to multitask learning of more than $2$ tasks by adding task-specific diagonal matrices, $M^{(1)}, \cdots, M^{(t)}$. For $t$ numbers of tasks, the diagonal matrix $M$ can be expanded and shared across all tasks as

\begin{equation}
  M :=  M^{(1)} \circ \cdots \circ M^{(t)}
  \label{eq:dot}
\end{equation}

where $M$ is computed from the matrices $ M^{(1)}, \cdots, M^{(t)}$ and $\circ$ is the Hadamard product or standard matrix product, which are the same for the diagonal matrices.

The overparameterisation and the parameter sharing mechanisms are shown in Figure \ref{fig:fac}.

\subsection{Iterative Training Strategy}
\label{sec:iter}

During the training phase of the overparameterised MTL model, instead of training the weight matrix $W$, the factorised matrices $U$, $M^{(1)}$ to $M^{(t)}$ and $V$ are trained. The $U$ and $V$ matrices are initialised using the same initialisation method as the original parameter matrix, and $M^{(1)}$ to $M^{(t)}$ are initialised into identity matrices. The trained weight matrices are contracted back to $W$ according to Eq. \eqref{eq:multiply} for inference with fewer parameter counts and computational costs.

In order to train shared and task-specific parameters separately, we propose an iterative training strategy which consists of two training processes for each epoch of training, as shown in Algorithm \ref{alg:training}

\begin{enumerate}[nolistsep]
    \item Choose a small subset of the training data ($\sim3\%$ in our experiments), for each task $j$ from all $t$ numbers of tasks to be learned together, only train its task-specific factors $M^{(j)}$ by task loss $L^{(j)}$, where $1 \leq j \leq t$. 
    \item All task-specific factors $M^{(0)}$ to $M^{(t)}$ are frozen. The other factors $U$, $V$ and the parameters of other unfactorised layers are trained by a multi-task learning loss $L=\sum_{i \in t} \alpha_iL^{(i)}$, where $\alpha$ represents the fixed or adaptive loss weights. 
\end{enumerate}
\begin{algorithm}[h]
    \caption{The Iterative Training Process}\label{alg:training}
    \begin{algorithmic}
    \Require Initialised parameters: $\theta$, loss weights $\alpha$
    \While{not convergence}
    \State sample a mini-batch data
    \For{$i\in\{1,\ldots,t\}$}
    \State update $M^{(i)}$ by $L^{(i)}$
    \EndFor
    \State update $\theta - \{M^{(1)},\ldots,M^{(t)}\}$ by $L=\sum_{i \in t} \alpha_iL^{(i)}$
    \EndWhile
    \end{algorithmic}
    \vspace{-0pt}
\end{algorithm}
During each training process, Frobenius decay is applied as a penalty on factorised matrices $U$, $V$ and $M$ to regularise the models for better generalisation, implemented as in \cite{khodak2021initialization}. The Frobenius decay rate is set to $1\text{e-}4$.

\section{Experiments}

Experiments are conducted on public datasets NYUv2 \cite{silberman2012indoor} and COCO \cite{lin2014microsoft} including a series of ablation studies and comparisons.  Note that as the proposed overparameterized model is contracted back to its original size for inference, the FLOPS, latency, and parameter size during inference following the proposed methods remain identical to the original model without overparameterization.

\subsection{Semantic Segmentation}
To first demonstrate the effectiveness of matrix factorisation as an overparameterisation method, we compare the method with other state-of-the-art overparameterisation methods, including RepVGG \cite{ding2021repvgg} and ExpandNet \cite{guo2020expandnets}, to perform a single Semantic Segmentation task. All experiments are conducted on NYUv2 dataset using SegNet model \cite{badrinarayanan2017segnet}, a smaller SegNet model (tagged with -S in the result tables) with all filter sizes for all convolutional layers (except for the final layer) cut into half. The overparameterisation is applied to all convolutional layers except for the final layer. The implementation of SegNet and all experiments on NYUv2 dataset follow the same training hyperparameters and configurations as in \cite{liu2019end}. The batch size is $8$. The total training epoch is $200$. A learning rate scheduler is used to reduce the learning rate to half at the $150$th epoch. Note that we use the multi-branch architecture including additional $1\times1$ convolutional and identity branches proposed in RepVGG to overparameterise the SegNet model, though it is not intended to be used as a drop-in replacement. Results in Table \ref{tab:ablation} show that the matrix factorisation method in this paper outperforms all other methods.


\begin{table}[htp]
\small
\centering
\resizebox{\linewidth}{!}{
\begin{tabular}{lccc}
\hline
\multicolumn{1}{c}{\multirow{2}{*}{\textbf{Method}}} &  & \multicolumn{2}{c}{\textbf{SemSeg}} \\ \cline{3-4} 
        &  & mIoU $\uparrow$& Pix Acc $\uparrow$          \\\hline
SegNet Baseline &&
 17.85 &
  55.36 
   \\ 
\textbf{Matrix Factorisation} &&
 \textbf{19.74} &
  \textbf{56.42}
   \\ 
RepVGG &&
 19.26&
  56.30 
   \\ 
ExpandNet &&
  16.27 &
  51.31 
   \\ 
SegNet-S Baseline &&
  17.22 &
  54.29
   \\ 
\textbf{SegNet-S Matrix Factorisation} &&
  \textbf{19.58} &
  \textbf{56.28} 
   \\ 
SegNet-S RepVGG &&
  17.25 &
  54.26 
   \\ 
SegNet-S ExpandNet &&
  16.59 &
  51.44 
   \\ \hline
\end{tabular}
}
\caption{Matrix Factorisation and SOTA Comparison on NYUv2}
\label{tab:ablation}
\end{table}

\begin{table}[b]
\centering
\resizebox{\linewidth}{!}{
\begin{tabular}{lcccllllllllll}
\hline
\multicolumn{1}{c}{\multirow{2}{*}{\textbf{Method}}} &  & \multicolumn{2}{c}{{\textbf{SemSeg}}} &  & \multicolumn{2}{c}{{\textbf{BBox}}}    &  & \multicolumn{2}{c}{\textbf{Instance Seg}}                    &  & \multicolumn{3}{c}{\textbf{Panoptic Seg}}                                                \\  \cline{3-4} \cline{6-7} \cline{9-10} \cline{12-14} 
\multicolumn{1}{c}{}                                 &  & mIoU $\uparrow$                    & Pix Acc $\uparrow$                 &  & \multicolumn{1}{c}{ AP $\uparrow$} & \multicolumn{1}{c}{mAP $\uparrow$} &  & \multicolumn{1}{c}{AP $\uparrow$} & \multicolumn{1}{c}{ mAP $\uparrow$} &  & \multicolumn{1}{c}{PQ $\uparrow$} & \multicolumn{1}{c}{SQ $\uparrow$} & \multicolumn{1}{c}{RQ $\uparrow$} \\ \hline

 Baseline &&
  21.34	& 69.15 &&
11.30 &	11.18 &&
  10.89 &	9.88	&&
  16.59	& 61.09	& 21.35
   \\ 
\textbf{Fac} &&
  \textbf{23.44} & \textbf{70.22} &&
  \textbf{15.62} & \textbf{16.11} &&
  \textbf{15.11} & \textbf{15.37} &&
  \textbf{20.94} & \textbf{64.40} & \textbf{26.64}
   \\ \hline
\end{tabular}}
\caption{Instance and Semantic Segmentation using PanopticFPN}
\label{tab:coco}
\end{table}

\subsection{Semantic Segmentation, Depth Estimation and Surface Normal Estimation}

\begin{table*}[!t]
\centering
\resizebox{0.9\linewidth}{!}{
\begin{tabular}{lccclllllllll}
\hline
\multicolumn{1}{c}{\multirow{3}{*}{\textbf{Method}}} &  & \multicolumn{2}{c}{\multirow{2}{*}{\textbf{SemSeg}}} &  & \multicolumn{2}{c}{\multirow{2}{*}{\textbf{Depth}}}    &  & \multicolumn{5}{c}{\textbf{Normal}}                                                                                                        \\ \cline{9-13} 
\multicolumn{1}{c}{}                                 &  & \multicolumn{2}{c}{}                                 &  & \multicolumn{2}{c}{}                                   &  & \multicolumn{2}{c}{Angel D}                            & \multicolumn{3}{c}{Within t}                                                      \\ \cline{3-4} \cline{6-7} \cline{9-13}

\multicolumn{1}{c}{}                                 &  & mIoU $\uparrow$                   & Pix Acc $\uparrow$              &  & \multicolumn{1}{c}{Abs Err $\downarrow$} & \multicolumn{1}{c}{Real Err $\downarrow$} &  & \multicolumn{1}{c}{ Mean $\downarrow$} & \multicolumn{1}{c}{Median $\downarrow$} & \multicolumn{1}{c}{11.25 $\uparrow$} & \multicolumn{1}{c}{22.5 $\uparrow$} & \multicolumn{1}{c}{30 $\uparrow$} \\

\hline
Baseline &&
 18.34 &
  55.97 & &
  0.5825 &
   0.2480 & &
  31.86 &
   26.83& 
   0.2033 & 
   0.4292 & 0.5529
   \\ 
Cross stitch &&
  18.68 &
  57.37 & &
  \textbf{0.5767} &
   \textbf{0.2499} & &
  31.68 &
   26.50 & 
   0.2098 & 
   0.4326 & 0.5560
   \\
\textbf{Fac} &&
\textbf{21.10} &
  \textbf{58.91} &&
  0.5784 &
  0.2419 &&
 \textbf{31.23} &
  \textbf{26.13} &
  \textbf{0.2151} &
  \textbf{0.4392} & \textbf{0.5610}
   \\
   Fac w/o iter &&
  20.37 &
  57.81 &&
  0.5836 &
  0.2482 &&
  31.35 &
  26.34&
  0.2149 &
  0.4363 & 0.5583
   \\ 
uvshare &&
  19.38 &
  57.61 &&
 0.5761 &
  0.2389 &&
  31.64 &
  26.71 &
  0.2078 &
  0.4284 & 0.5524
   \\
mshare &&
  18.55 &
  55.33 &&
  0.5961 &
  0.2539 &&
  32.35 &
  27.30 &
  0.2037 &
  0.4215 & 0.5422
   \\ \hline
Baseline-S &&
  16.43 &
  53.07 & &
  0.6332 &
   0.2664 & &
  34.88 &
   29.89& 
   0.1865 & 
   0.3873 & 0.5019
   \\ 
Cross stitch-S &&
  18.07 &
  55.47 & &
  \textbf{0.6040} &
   \textbf{0.2502} & &
  \textbf{32.83} &
   \textbf{27.63} & 
   \textbf{0.2071} & 
   \textbf{0.4184} & \textbf{0.5368}
   \\ 
\textbf{Fac-S} &&
  \textbf{18.46}&	\textbf{55.70}&&
0.6080&	0.2634&&
  33.35	&28.71	&0.1915	&0.4002	&0.5199
   \\ 
Fac-S w/o iter &&
  18.13&	55.97&&
0.6069&	0.2577&&
  34.16	&30.00	&0.1833	&0.3830	&0.5005
   \\ \hline
   MTAN Baseline &&
  19.40	& 56.67 &&
0.6074 &	0.2506 &&
  31.41 &	27.57	&0.1858	&0.4120	&0.5440
   \\ 
\textbf{MTAN Fac} &&
  \textbf{21.12} &
  \textbf{57.89} &&
  \textbf{0.6023} &
  \textbf{0.2491} &&
  \textbf{30.23}  &
  \textbf{25.49}	& \textbf{0.2204} & \textbf{0.4499} & \textbf{0.5764}
   \\ \hline
   \hline
Baseline-DeepLabv3 &&
 46.13 &
  75.83 & &
  0.4236 &
   0.1633 & &
    23.74 &
   16.42 & 
   0.3701 & 
   0.6231 & 0.7227
   \\ 
Fac-DeepLabv3 w/o iter &&
  46.84 &
  \textbf{80.35} & &
  \textbf{0.4089} &
   0.1606 & &
  23.15 &
   15.92 & 
   0.3818& 
   0.6342 & 0.7325
   \\ 
\textbf{Fac-DeepLabv3} &&
\textbf{47.14} &
  80.25 &&
  0.4095 &
  \textbf{0.1581} &&
 \textbf{23.03} &
  \textbf{15.83} &
  \textbf{0.3841} &
  \textbf{0.6372} & \textbf{0.7348} \\
   \hline 
Baseline-DeepLabv3 Pretrained  &&
 \textbf{55.76} &
  89.24 & &
  0.3474 &
   0.1378 & &
    \textbf{21.34} &
   14.74 & 
   0.4039 & 
   \textbf{0.6687} & \textbf{0.7671}
   \\ 
Fac-DeepLabv3 w/o iter Pretrained &&
  55.05 &
  86.51 & &
  0.3463 &
   0.1356 & &
  21.55 &
   14.85 & 
   0.4030 & 
   0.6639 & 0.7623
   \\ 
\textbf{Fac-DeepLabv3 Pretrained} &&
55.56 &
  \textbf{90.10} &&
  \textbf{0.3432} &
  \textbf{0.1356} &&
 21.46 &
  \textbf{14.70} &
  \textbf{0.4064} &
  0.6678 & 0.7647 \\
   \hline 
\end{tabular}}
\caption{Performance Comparison and Ablation Studies with SegNet and DeepLabv3}
\label{tab:segnet}
\end{table*}

The proposed parameter-sharing mechanisms and iterative training strategy for MTL are experimented on NYUv2 dataset using SegNet, small SegNet and DeepLabv3 \cite{chen2017rethinking} with ResNet50 backbone (tagged with -ResNet). 
For some experiments (tagged with Pretrain), the ResNet50 backbone is pre-trained with ImageNet 1k dataset. All convolutional layers in the shared backbone are overparameterised for SegNet models. For DeepLabv3 model, all the first convolutional layers in ResNet bottleneck modules are overparameterised following the similar design in YOLOv7 \cite{wang2022yolov7}. 
The initialisation of the overparameterised parameters follows the spectral initialisation introduced in \cite{khodak2021initialization}. The performance is compared with multiple MTL methods including MTL baseline with only shared backbones and task-specific heads, cross-stitch \cite{misra2016cross} and MTAN \cite{liu2019end}. The batch size for MTAN and the proposed method applied to MTAN is set to $2$, and learning rate is $5\text{e-}5$. 
Note that the task-specific diagonal matrices are trained without weight decay in the optimiser.

Ablation studies are also conducted to compare the proposed method (\textbf{Fac}) with a series of variants:

\begin{itemize}[nolistsep]
    \item \textbf{Fac w/o iter} - The shared backbone is overparameterised according to Figure \ref{fig:fac}, but all factors are trained with the combined task losses.
    \item \textbf{uvshare} - Split the channels factorised in factors $U$ and $V$ evenly across tasks, and stack the split matrix back to recover $U$ and $V$.
    \item \textbf{mshare} - Split the core diagonal matrix $M$ evenly across tasks, and stack the split matrix back to recover $M$.
\end{itemize}

Experiment results are shown in Table \ref{tab:segnet}. The proposed method outperforms all baselines except for some tasks under the cross-stitch model, which has increased expressiveness by enlarging the model size ($\sim3$ times larger in inference model parameter size). The proposed method showcases comparable or superior performance in comparison to a backbone model pretrained with the ImageNet dataset, even when the backbone structure experiences changes due to overparameterisation, resulting in the loss of the pretrained weights

\subsection{Instance and Semantic Segmentation}

We also evaluate the performance of our proposed method using Panoptic Feature Pyramid Network (PanopticFPN) \cite{kirillov2019panoptic} to perform instance segmentation and semantic segmentation tasks simultaneously. We follow the same experiment configurations described in \cite{kirillov2019panoptic} using COCO dataset. By overparameterise the PanopticFPN and following the iterative training strategy, our model also performs better than the Baseline model without pretraining on other datasets, as shown in Table \ref{tab:coco}.

\section{Related Works}

After investigating the advantages of optimisation in overparameterised linear and nonlinear networks in ~\cite{arora2018optimization}, several studies have proposed overparameterised deep learning models based on fully-connected or convolutional neural networks. However, none of these studies designed their overparameterisation under the context of MTL, accounting for the effective parameter sharing among multiple tasks. Among these studies, 
ExpandNet~\cite{guo2020expandnets}, DO-Conv~\cite{cao2020conv} and Overcomplete Knowledge Distillation~\cite{khodak2021initialization} designed different methods to overparameterise the kernels of convolutional layers, either over their channel axes~\cite{guo2020expandnets} or spatial axes~\cite{cao2020conv}. A more recent design named RepVGG~\cite{ding2021repvgg} is a special case of overparameterisation, which instead of designing a drop-in replacement for convolutional layers of any neural architecture as previous studies, presented an effective convolutional neural network model with an overparameterised training network and a VGG-like inference network.

\section{Conclusion}

In this paper, we propose a parameter-sharing scheme and an iterative training for deep multitask learning that effectively share parameters using overparameterised models during training, while the model architecture stayed slim and compact during inference. Compared to the state-of-the-art, the scheme has demonstrated its potential in various datasets and model architecture. The design is particularly well-suited for embedded computer vision applications where there are tight budget constraints in both memory and computational resources.


\section*{Acknowledgements}

This study is supported under the RIE2020 Industry Alignment Fund - Industry Collaboration Projects (IAF-ICP) Funding Initiative, as well as cash and in-kind contribution from the industry partner(s).

\bibliographystyle{named}
\bibliography{egbib}

\end{document}